\documentclass[letterpaper,10pt,conference]{ieeeconf}

\IEEEoverridecommandlockouts
\usepackage{graphicx}
\usepackage{bm}
\usepackage{amsmath}
\usepackage{amssymb}
\usepackage{amsfonts}
\usepackage{booktabs}
\usepackage{array}
\usepackage{makecell}
\usepackage{multirow}
\usepackage{cite}
\usepackage{url}
\usepackage{capt-of}
\usepackage{caption}
\usepackage{subcaption}
\usepackage{cuted}
\usepackage{xcolor}
\usepackage[hidelinks]{hyperref}
\definecolor{urlcolor}{RGB}{237,2,140}

\title{Stay Seated: Learning Omnidirectional Humanoid Locomotion \\ on a Passive Mobile Chair with Casters}

\author{Kango Yanagida$^{1}$, Kazuki Miyazawa$^{1}$, and Takato Horii$^{1,2}$%
\thanks{*This work was partially supported by JST Moonshot R\&D Program, Japan, under Grant Numbers JPMJMS2011 and JPMJMS263D, and by JST BOOST, Japan, under Grant Number JPMJBS2402.}%
\thanks{$^{1}$Dept. of Systems Innovation, Graduate School of Engineering Science, The University of Osaka, Japan.}%
\thanks{$^{2}$International Research Center for Neurointelligence, The University of Tokyo, Japan.}%
\thanks{\ttfamily\footnotesize \{yanagida.kango.z7m@ecs., miyazawa.kazuki.es@, takato@sys.es.\} osaka-u.ac.jp}
\\{\normalsize\color{urlcolor}\url{https://kan5r.github.io/stay-seated/}}
}%

\begin{document}
\maketitle
\thispagestyle{empty}
\pagestyle{empty}

\begin{abstract}
Humanoid robots with quasi-direct-drive actuators continuously generate joint torque while standing, whereas seated humans delegate weight support to chairs during desk work. As a first step toward seated loco-manipulation, we study omnidirectional seated locomotion on a passive mobile chair with casters, a robot--chair--floor system that requires unfixed pelvis--seat contact and intermittent foot--floor propulsion of the robot and chair. We ask whether this task can be learned by minimally extending a standard standing velocity-tracking environment, adding only a passive-chair model, seated-state rewards, critic-only chair observations, and pelvis--seat contact settings. The policy is learned without motion references, and its actor uses only proprioception and velocity commands. In random-command evaluation, the policies tracked omnidirectional commands through nearly all 20-s rollouts with accuracy comparable to a standing locomotion policy. Comparing symmetry regularization (SY), foot-slip penalty (FS), and command curriculum (CC) showed that FS reduced the cost of transport (CoT) but increased tracking error, and that some FS-only policies converged to stationary local optima under diagonal-forward commands. Adding either SY or CC avoided these local optima. CC first learns low-speed propulsion, where smaller foot forces suffice to propel without slipping, and then expands the command range. SY further reduced the left--right bias in leg use that otherwise persisted, likely because the chair supports the body. Direction-resolved analysis showed CoT ordered backward $<$ lateral $\ll$ forward, with planted-leg extension in backward and lateral motion and knee flexion following heel contact in forward motion. The learned policy achieved zero-shot sim-to-real transfer to a Unitree G1.
\end{abstract}

\section{Introduction}
\label{sec:intro}
Most humanoid locomotion tasks based on deep reinforcement learning (DRL) assume a standing posture.
Humanoids equipped with quasi-direct-drive (QDD) motors must continuously generate joint torque to support their weight while standing still, resulting in nonzero holding current and heat generation.
Humans, in contrast, typically perform desk work while seated and delegate body-weight support to a chair. A chair with casters also permits changes in working position and viewpoint without standing up. 
Seated behavior can therefore complement standing behavior during extended tabletop tasks.
This motivates our goal of realizing \emph{seated loco-manipulation}, in which a humanoid moves and manipulates objects while remaining seated. 
As a first step, this study addresses \emph{seated locomotion} on a passive mobile chair. The robot tracks target translational velocity $(v_x,v_y)$ and yaw rate $\omega_z$ while remaining seated.

Prior work has applied DRL to humanoid locomotion with wheeled and sliding platforms~\cite{skater,marot2026inline,thibault2024skateboarding,han2026husky,wu2021ski,baltes2023scooter}, in which body--device contact is typically constrained in advance and ground interaction occurs mainly through the device. Seated locomotion instead forms a robot--chair--floor system in which unfixed pelvis--seat contact is combined with direct foot--floor propulsion of the robot and chair, a configuration that, to our knowledge, DRL-based humanoid locomotion has not addressed (Sec.~\ref{chap:related_work}). We ask whether it can be acquired by minimally extending a standard standing velocity-tracking environment, without a dedicated motion reference or a new learning algorithm. 

\begin{figure}[t]
    \centering
    \includegraphics[width=\linewidth]{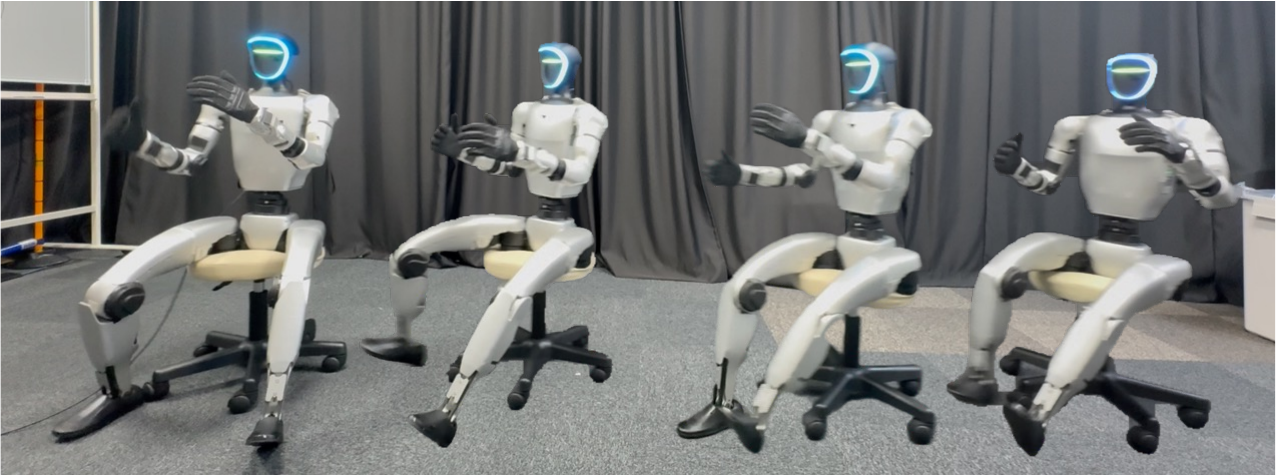}
    \caption{Seated locomotion on a passive mobile chair with casters during translation and turning. }
    \label{fig:real_robot_teaser}
\end{figure}

The contributions of this study are as follows:
\begin{enumerate}
\item We formulate seated locomotion on a passive mobile chair as an omnidirectional velocity tracking task with unfixed pelvis--seat contact and intermittent foot--floor propulsion. We show that minimal extensions to a standard standing velocity-tracking environment enable learning without motion references, and that an actor using only proprioception and velocity commands transfers zero-shot to a physical Unitree G1.
\item We show that a command curriculum (CC), which learns low-speed propulsion before expanding the command range, avoids the stationary solutions to which some training runs otherwise converge under large diagonal-forward commands. These solutions likely arise because body weight is shared between the chair and the feet, so propulsion requires pressing the feet against the floor without slipping.
\item We also show that symmetry regularization (SY)~\cite{mittal2024symmetry} reduces the left--right bias in leg use, balancing left and right contact fractions and forces. This bias likely persists because the chair supports the body, so biased leg use barely affects stability or velocity tracking, whereas in standing locomotion it compromises posture and is corrected during learning.
\end{enumerate}

\section{Related Work}
\label{chap:related_work}
\subsection{Humanoid Locomotion with Wheeled and Sliding Platforms}

Wheeled and sliding platforms have been studied as a means of extending humanoid mobility, including roller and inline skates~\cite{skater,marot2026inline}, skateboards~\cite{thibault2024skateboarding,han2026husky}, skis~\cite{wu2021ski}, and two-wheeled scooters~\cite{baltes2023scooter}.
These studies achieve dynamic locomotion while accounting for the coupled dynamics of the humanoid and mobility device, as well as nonholonomic constraints arising from wheels or sliding surfaces.

In many existing tasks, the mobility device is attached to the feet or the body--device contact relationship is constrained in advance, and interaction with the ground occurs primarily through the device~\cite{skater,marot2026inline,baltes2023scooter,wu2021ski}.
Skateboarding is an exception involving direct foot--ground propulsion~\cite{thibault2024skateboarding,han2026husky}; however, pushing against the ground and steering with both feet on the board are treated as distinct motion phases in HUSKY~\cite{han2026husky}.

Seated locomotion instead requires the robot to propel itself and a passive chair directly through foot--floor contact while maintaining unfixed pelvis--seat contact and tracking omnidirectional translational and yaw commands.
To the best of our knowledge, DRL-based humanoid locomotion has not jointly addressed unfixed body--tool contact, direct foot--floor propulsion, and omnidirectional velocity-command tracking.
We do not rank this task against prior wheeled and sliding locomotion by difficulty; it has a different contact topology and control objective.

Kawaharazuka et al.~\cite{kawaharazuka} demonstrated forward, backward, and turning seated walking on a tendon-driven musculoskeletal humanoid through human-constrained teaching.
In contrast, we target a QDD humanoid and learn omnidirectional seated locomotion without motion references using task rewards and regularization terms.
The actor uses only proprioception and velocity commands and receives neither foot--floor nor pelvis--seat contact sensing nor chair states.

\subsection{Reference-Based Humanoid Locomotion Learning}

Reference-based DRL methods, including BeyondMimic~\cite{liao2025beyondmimic} and AMP~\cite{peng2021amp}, use human motion-capture data as reference motion to acquire natural and diverse whole-body motions.
For seated locomotion, however, the human and chair motions and their contact states would have to be captured synchronously and retargeted for various directions and speeds.
Recent retargeting methods preserve kinematic interaction constraints~\cite{yang2025omniretarget} or enforce dynamic feasibility for foot--floor~\cite{zhang2026kinodynamic} or body--object~\cite{dhedin2026dynaretarget} contacts; seated locomotion requires both contact types to be reconciled dynamically at the same time, leaving reference-motion generation complex.

Analytical gait generators such as CPGs~\cite{nassour2014multilayered,cristiano2017generation} have primarily addressed standing locomotion, and appropriate contact cycles for omnidirectional seated locomotion remain unclear.

We therefore learn the leg coordination required for seated locomotion using only task-specific rewards and regularization terms, without reference motions or gait templates.

\section{Method}

\begin{figure}[t]
    \centering
    \includegraphics[width=\linewidth]{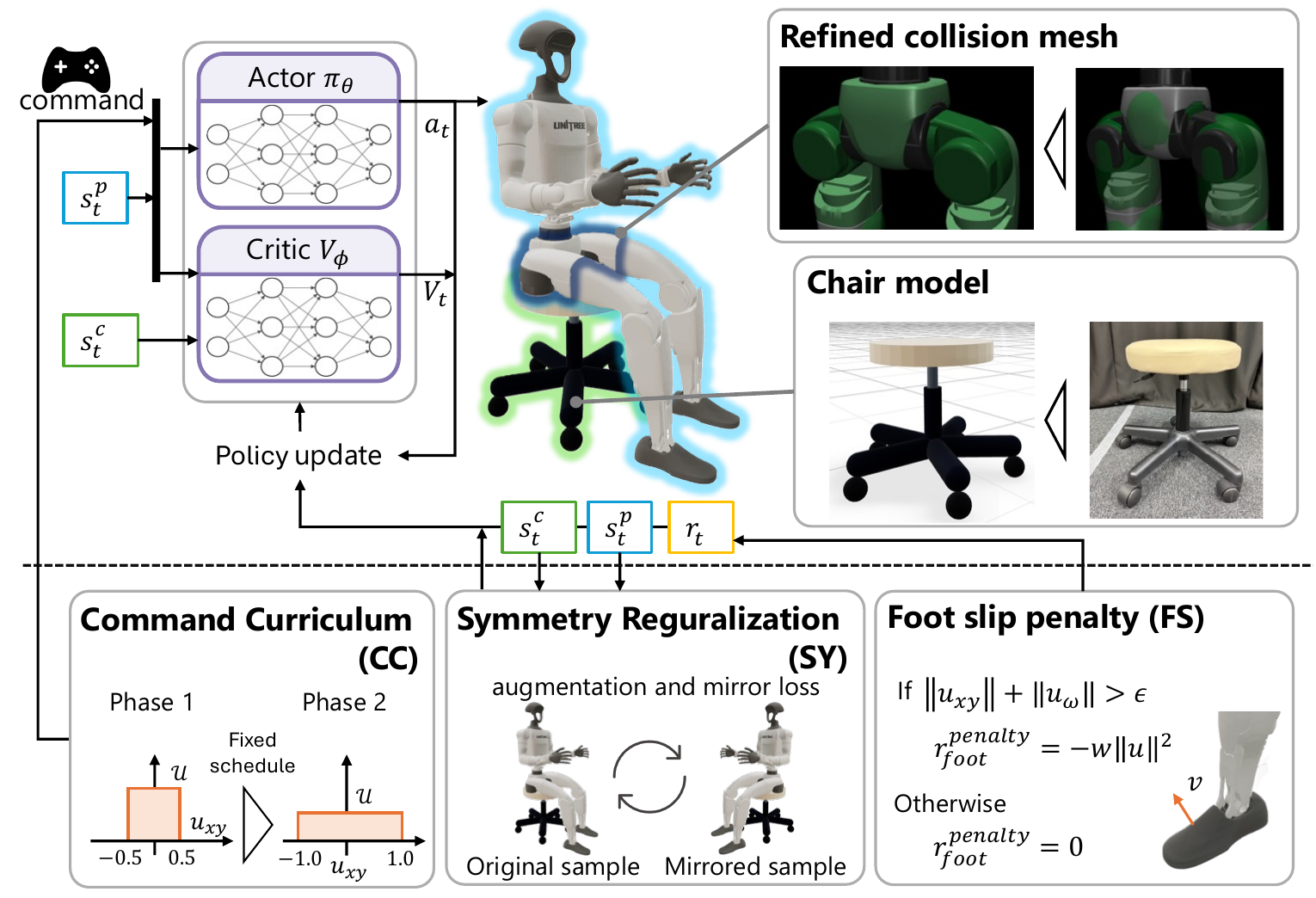}
    \caption{Overview of the seated locomotion learning framework, including the asymmetric actor--critic architecture, refined pelvis collision mesh, and passive-chair model. The full-factorial study varies command curriculum (CC), symmetry regularization (SY), and foot-slip penalty (FS).}
    \label{fig:rl_framework}
\end{figure}

Starting from a standard standing velocity-tracking environment, we introduce the changes required for seated locomotion: a passive-chair model, rewards describing the seated state, chair-related privileged states provided only to the critic, and contact settings for pelvis--seat interaction.
The overall learning framework, simulation models, and three training components are summarized in Fig.~\ref{fig:rl_framework}.
This section describes the seated-locomotion problem, robot--chair--floor system, observation and action spaces, reward, and training conditions.
We use mjlab~\cite{zakka2026mjlab} as the simulation training environment.

\subsection{Seated-Locomotion Problem Formulation}

We formulate seated-locomotion control as a discrete-time Markov decision process $\mathcal{M}=(\mathcal{S},\mathcal{A},P,r,\gamma)$.
The state $s_t$ comprises the simulated states of the robot and chair, their contact states, and the velocity command.

At each time step $t$, the actor receives an observation $\bm{o}^{\mathrm{actor}}_t$ constructed from $s_t$ and samples an action according to $\bm{a}_t\sim\pi_\phi(\bm{a}_t\mid\bm{o}^{\mathrm{actor}}_t)$.
The policy parameters $\phi$ are optimized to maximize the expected discounted return
\begin{equation}
    J(\phi)
    =
    \mathbb{E}_{\pi_\phi}
    \left[
        \sum_{t=0}^{T-1}
        \gamma^t r_t
    \right].
\end{equation}

The policy is conditioned on the torso-frame forward and lateral translational velocities and yaw rate $\bm{u}_t=[v^{\mathrm{cmd}}_{x,t},v^{\mathrm{cmd}}_{y,t},\omega^{\mathrm{cmd}}_{z,t}]^{\mathsf T}$.
Without rigidly attaching the robot to the chair, the policy must maintain pelvis--seat contact and track $\bm{u}_t$ while propelling both bodies through intermittent foot--floor contact.

\subsection{Robot--Chair--Floor System and Contact Modeling}

The system comprises a 29-degree-of-freedom Unitree G1 and a mobile chair with five passive casters and a rotating seat.
The physical chair and its simulation model are shown in the chair-model inset of Fig.~\ref{fig:rl_framework}.
Each caster is modeled as a sphere connected to a chair leg through a ball joint.
The translational command ranges are $v_x^{\mathrm{cmd}},v_y^{\mathrm{cmd}}\in[-1.0,1.0]$~m/s, and the yaw-velocity command range is $\omega_z^{\mathrm{cmd}}\in[-0.5,0.5]$~rad/s.
Each episode lasts 20~s, corresponding to 1,000 control steps.
An episode terminates early if the torso or chair tilts by more than $70^{\circ}$ from the gravity direction or if pelvis--seat contact is lost continuously for at least 1~s.

Because contact between the pelvis and seat strongly affects the task, we refine the collision mesh around the pelvis to match the visual mesh, as shown in the refined-collision-mesh inset of Fig.~\ref{fig:rl_framework}.
We also change the friction cone used by the contact solver from a pyramidal cone to an elliptic cone.
The simulation time step is 0.005~s with a control decimation of four.
For the MuJoCo solver, we use \texttt{impratio}=10, 15 solver iterations, 20 line-search iterations, and 50 continuous-collision-detection iterations.
These settings suppress excessive pelvis--seat penetration and unnatural lateral slipping observed with the coarse pelvis geometry and default contact settings.

\subsection{Observation and Action Spaces}
\label{subsec:observation_action}

\begin{table}[t]
    \centering
    \caption{Observation and action spaces for the seated locomotion task.}
    \label{tab:task_components}
    \scriptsize
    \setlength{\tabcolsep}{3pt}
    \begin{tabular}{@{}
                    >{\raggedright\arraybackslash}p{0.19\columnwidth}
                    >{\raggedright\arraybackslash}p{0.50\columnwidth}
                    r@{}}
        \toprule
        Type & Component & Space \\
        \midrule
        \multirow{5}{*}{\makecell[l]{Actor and\\critic\\observations}}
            & Torso angular velocity & $\mathbb{R}^{3}$ \\
            & Projected gravity & $\mathbb{R}^{3}$ \\
            & Joint positions / velocities & $\mathbb{R}^{29} / \mathbb{R}^{29}$ \\
            & Last action & $\mathbb{R}^{29}$ \\
            & Velocity command & $\mathbb{R}^{3}$ \\
        \midrule
        \multirow{9}{*}{\makecell[l]{Critic-only\\observation}}
            & Torso linear velocity & $\mathbb{R}^{3}$ \\
            & Foot heights & $\mathbb{R}^{2}$ \\
            & Foot air times & $\mathbb{R}^{2}$ \\
            & Foot contact states / forces & $\mathbb{R}^{2} / \mathbb{R}^{6}$ \\
            & Chair-relative pose / velocity & $\mathbb{R}^{9} / \mathbb{R}^{6}$ \\
            & Chair projected gravity & $\mathbb{R}^{3}$ \\
            & Caster contact states / forces & $\mathbb{R}^{5} / \mathbb{R}^{15}$ \\
            & Pelvis--seat contact states / forces & $\mathbb{R}^{3} / \mathbb{R}^{9}$ \\
            & Chair--leg contact states / forces & $\mathbb{R}^{14} / \mathbb{R}^{42}$ \\
        \midrule
        Action
            & Normalized joint-position command & $\mathbb{R}^{29}$ \\
        \bottomrule
    \end{tabular}
\end{table}

To facilitate sim-to-real transfer, we adopt an asymmetric actor--critic architecture~\cite{pinto2018asymmetric}.
As shown in the actor--critic block of Fig.~\ref{fig:rl_framework}, the actor observation $\bm{o}^{\mathrm{actor}}_t\in\mathbb{R}^{96}$ contains only proprioceptive quantities available on hardware and the velocity command; it excludes pelvis--seat contact, foot--floor contact, and chair states.
During training, the critic receives the uncorrupted actor observation together with simulator-only robot, chair, and contact information, yielding $\bm{o}^{\mathrm{critic}}_t\in\mathbb{R}^{217}$.
Table~\ref{tab:task_components} gives the complete composition.

The actor outputs a normalized joint-position command $\bm{a}_t\in\mathbb{R}^{29}$ centered on the initial seated posture.
It is converted to the proportional--derivative controller target as $\bm{q}_{\mathrm{target},t}=\bm{q}_{\mathrm{default}}+\bm{s}\odot\bm{a}_t$, where $\bm{q}_{\mathrm{default}}$ denotes the joint positions of the initial seated posture, $\bm{s}$ is a joint-wise action scale, and $\odot$ denotes element-wise multiplication.

\begin{table}[t]
    \centering
    \begin{minipage}[t]{0.41\columnwidth}
        \centering
        \captionsetup{justification=centering}
        \captionof{table}{Network and training settings.}
        \label{tab:training_hyperparameters}
        \scriptsize
        \setlength{\tabcolsep}{0pt}
        \begin{tabular}{@{}lr@{}}
            \toprule
            Parameter & Value \\
            \midrule
            \makecell[l]{MLP hidden\\layers} & 512,256,128 \\
            Activation & ELU \\
            Learning rate & $1.0\times10^{-3}$ \\
            Discount factor $\gamma$ & 0.99 \\
            GAE parameter $\lambda$ & 0.95 \\
            Clipping range $\epsilon$ & 0.2 \\
            Entropy coefficient & 0.01 \\
            Control frequency & 50 Hz \\
            Parallel environments & 4,096 \\
            PPO iterations & 10,000 \\
            Random seeds & \makecell[r]{42, 1234,\\2026, 3407} \\
            Rollout horizon & 24 steps \\
            \bottomrule
        \end{tabular}
    \end{minipage}
    \hfill
    \begin{minipage}[t]{0.57\columnwidth}
        \centering
        \captionsetup{justification=centering}
        \captionof{table}{Domain randomization ranges.}
        \label{tab:domain_randomization}
        \scriptsize
        \setlength{\tabcolsep}{1.0pt}
        \begin{tabular}{@{}lr@{}}
            \toprule
            Component & Range \\
            \midrule
            \makecell[l]{Linear velocity}
                & \makecell[r]{$x,y:\pm0.2$ m/s\\$z:\pm0.1$ m/s} \\
            \makecell[l]{Angular velocity}
                & \makecell[r]{roll, pitch: $\pm0.2$ rad/s\\yaw: $\pm0.3$ rad/s} \\
            Foot friction & $[0.3,1.2]$ \\
            Joint-encoder bias & $\pm0.015$ rad \\
            Torso CoM
                & \makecell[r]{$x,y:\pm0.025$ m\\$z:\pm0.03$ m} \\
            \makecell[l]{Chair-base CoM}
                & \makecell[r]{$x,y:\pm0.02$ m\\$z:\pm0.01$ m} \\
            Chair-base height & $\pm0.02$ m \\
            Caster position & $\pm0.015$ m \\
            Chair-base friction & $[0.8,1.4]$ \\
            \makecell[l]{Caster damping} & $[0.6,1.4]$ \\
            Chair-hinge damping & $[1.4,2.2]$ \\
            \makecell[l]{Projected-gravity} & $\pm0.05$ \\
            Joint position & $\pm0.01$ rad \\
            Joint velocity & $\pm1.5$ rad/s \\
            \bottomrule
        \end{tabular}
    \end{minipage}
\end{table}

\subsection{Reward Design}
\label{subsec:reward_design}

The reward comprises terms for velocity-command tracking, posture stabilization, maintenance of the seated state, foot motion, and motion regularization.
The reward at each time step is
\begin{equation}
    r
    =
    r_{\mathrm{task}}
    +
    r_{\mathrm{posture}}
    +
    r_{\mathrm{seat}}
    +
    r_{\mathrm{foot}}^{\mathrm{penalty}}
    +
    r_{\mathrm{reg}}^{\mathrm{penalty}}.
\end{equation}
Here, $r_{\mathrm{foot}}^{\mathrm{penalty}}\leq0$ and $r_{\mathrm{reg}}^{\mathrm{penalty}}\leq0$ are penalties for foot motion and motion regularization, respectively.

The task reward $r_{\mathrm{task}}$ encourages tracking of the target translational and yaw velocities.
The posture reward $r_{\mathrm{posture}}$ comprises rewards for keeping the torso and chair upright and for limiting joint deviation from the initial seated posture.

The seated-state reward $r_{\mathrm{seat}}$ uses the pelvis position $\bm{p}_{\mathrm{pelvis}}$, chair position $\bm{p}_{\mathrm{chair}}$, and pelvis--seat contact state:
\begin{equation}
    \begin{split}
        r_{\mathrm{seat}} ={}&
        w_{\mathrm{seat,pos}}
        \exp\left(
            -\frac{
                \left\|
                    \bm{p}_{\mathrm{pelvis},xy}
                    -
                    \bm{p}_{\mathrm{chair},xy}
                \right\|_2^2
            }{
                \sigma_{\mathrm{seat}}^2
            }
        \right) \\
        &+
        w_{\mathrm{seat,contact}}
        \mathbb{I}_{\mathrm{contact}}.
    \end{split}
    \raisetag{10pt}
\end{equation}
Here and below, $\mathbb{I}(\cdot)$ denotes the indicator function, $\mathbb{I}_{\mathrm{contact}}$ indicates pelvis--seat contact, $w_{\mathrm{seat,pos}}=w_{\mathrm{seat,contact}}=0.5$, and $\sigma_{\mathrm{seat}}=0.20$~m.
Chair velocity is not directly tracked by a reward; instead, the chair is moved through robot velocity tracking and robot--chair contact.

The foot penalty $r_{\mathrm{foot}}^{\mathrm{penalty}}$ contains terms for foot clearance, tangential foot velocity during contact, and landing impact.
The factorial comparison uses the following slip penalty $r_{\mathrm{slip}}^{\mathrm{penalty}}$ with $\epsilon_{\mathrm{slip}}=0.3$:
\begin{equation}
    \begin{aligned}
        r_{\mathrm{slip}}^{\mathrm{penalty}}
        ={}&
        -w_{\mathrm{slip}}
        \mathbb{I}\left(\|\bm{u}_{xy,t}\|_2+|u_{\omega,t}|>\epsilon_{\mathrm{slip}}\right)
        \\
        &{}\times
        \sum_{i\in\mathcal{F}}
        \mathbb{I}^{i}_{\mathrm{contact},t}
        \left\|
            \bm{v}^{i}_{\mathrm{foot},xy,t}
        \right\|_2^2.
    \end{aligned}
\end{equation}
Here, $\bm{u}_{xy,t}$ and $u_{\omega,t}$ are the translational and yaw components of the command $\bm{u}_t$, $\mathcal{F}$ is the set of feet, $\mathbb{I}^{i}_{\mathrm{contact},t}$ indicates floor contact of foot $i$, and $\bm{v}^{i}_{\mathrm{foot},xy,t}$ is its planar velocity. The first indicator disables the penalty when the command magnitude is below $\epsilon_{\mathrm{slip}}$.

The regularization penalty $r_{\mathrm{reg}}^{\mathrm{penalty}}$ limits torso angular velocity, whole-body angular momentum, self-collision, joint-limit violation, joint acceleration, action rate, action acceleration, and deviation from the initial seated posture under a zero command.

\begin{table*}[!t]
    \centering
    \caption{Random-command evaluation.
        Values are the mean and standard deviation across four training seeds;
        each learned policy was evaluated over 1,000 rollouts.
        In each column, bold and underlined values denote the best and
        second-best values among the eight training conditions, respectively.}
    \label{tab:random_command_results}
    \scriptsize
    \setlength{\tabcolsep}{3.2pt}
    \begin{tabular}{lcccccc}
        \toprule
        Condition
            & \makecell{Tracking\\RMSE $v_x$\\{[m/s]}}
            & \makecell{Tracking\\RMSE $v_y$\\{[m/s]}}
            & \makecell{Tracking\\RMSE $\omega_z$\\{[rad/s]}}
            & \makecell{Timeout\\success [\%]}
            & \makecell{Seat-relative\\RMS [m]}
            & \makecell{Torso-tilt\\RMS [rad]} \\
        \midrule
        Baseline
            & $0.1650\pm0.0084$ & $0.1441\pm0.0071$ & $0.1506\pm0.0014$
            & $99.45\pm0.17$ & $0.0153\pm0.0008$ & $0.0459\pm0.0018$ \\
        SY
            & $0.1523\pm0.0043$ & $\underline{0.1284\pm0.0033}$ & $0.1444\pm0.0006$
            & $99.68\pm0.52$ & $0.0157\pm0.0008$ & $\underline{0.0450\pm0.0019}$ \\
        FS
            & $0.1964\pm0.0432$ & $0.1879\pm0.0473$
            & $\mathbf{0.1411\pm0.0099}$
            & $99.53\pm0.17$ & $\mathbf{0.0151\pm0.0009}$
            & $0.0480\pm0.0038$ \\
        CC
            & $0.1524\pm0.0023$ & $0.1341\pm0.0024$ & $0.1498\pm0.0021$
            & $99.73\pm0.17$ & $\underline{0.0152\pm0.0011}$ & $0.0456\pm0.0025$ \\
        SY+FS
            & $0.1542\pm0.0028$ & $0.1382\pm0.0033$ & $0.1457\pm0.0028$
            & $99.68\pm0.17$ & $0.0159\pm0.0012$ & $0.0469\pm0.0035$ \\
        SY+CC
            & $\mathbf{0.1512\pm0.0002}$
            & $\mathbf{0.1268\pm0.0036}$
            & $0.1453\pm0.0025$
            & $\mathbf{99.80\pm0.14}$
            & $0.0156\pm0.0023$
            & $\mathbf{0.0436\pm0.0030}$ \\
        FS+CC
            & $0.1562\pm0.0017$ & $0.1335\pm0.0032$ & $0.1501\pm0.0026$
            & $\underline{99.78\pm0.10}$ & $0.0156\pm0.0009$ & $0.0462\pm0.0013$ \\
        SY+FS+CC
            & $\underline{0.1522\pm0.0040}$ & $0.1292\pm0.0016$
            & $\underline{0.1442\pm0.0006}$ & $99.75\pm0.13$
            & $0.0156\pm0.0004$ & $0.0452\pm0.0011$ \\
        Standing
            & $0.1551\pm0.0018$ & $0.1490\pm0.0013$ & $0.1752\pm0.0056$
            & $99.70\pm0.36$ & -- & $0.0505\pm0.0002$ \\
        \bottomrule
    \end{tabular}
\end{table*}

\begin{figure*}[!t]
    \centering
    \includegraphics[width=\linewidth]{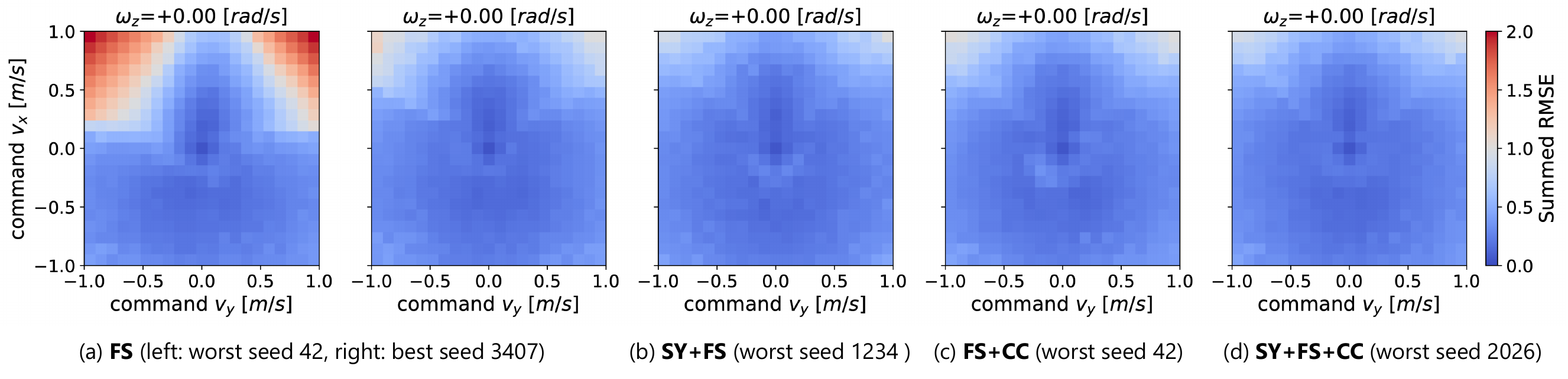}
    \caption{Translational tracking error heatmaps from the fixed-command evaluation for FS, SY+FS, FS+CC and SY+FS+CC.}
    \label{fig:tracking_error_heatmap}
\end{figure*}

\subsection{Training Setup and Factorial Conditions}
\label{subsec:training_conditions}

Each condition is trained independently with PPO~\cite{schulman2017ppo} in mjlab~\cite{zakka2026mjlab} using the settings in Table~\ref{tab:training_hyperparameters}.
For sim-to-real transfer, we use the domain randomization in Table~\ref{tab:domain_randomization}: model parameters are sampled at environment startup, velocity disturbances are applied every 2--5~s, and observation noise is added only to actor inputs.

The full-factorial comparison varies the three training components illustrated in the lower part of Fig.~\ref{fig:rl_framework}.
1) \emph{Symmetry regularization} (SY) jointly applies left--right symmetry-based data augmentation and adds a mirror-loss term to the actor loss with coefficient $\lambda_{\mathrm{mir}}=1.0$~\cite{mittal2024symmetry}. 
2) \emph{Foot-slip penalty} (FS) is the penalty on tangential foot velocity during contact defined in Sec.~\ref{subsec:reward_design}; $w_{\mathrm{slip}}=0$ disables FS, whereas $w_{\mathrm{slip}}=0.25$ enables it.
3) \emph{Command curriculum} (CC) uses a manually specified schedule for the translational command range.
For the first 2,000 PPO iterations, $v_x^{\mathrm{cmd}},v_y^{\mathrm{cmd}}\in[-0.5,0.5]$~m/s; thereafter, the range expands to $[-1.0,1.0]$~m/s.
The yaw-rate range remains $\omega_z^{\mathrm{cmd}}\in[-0.5,0.5]$~rad/s throughout training.

We compare the $2^3=8$ conditions obtained from the full factorial combination of SY, FS, and CC: Baseline, SY, FS, CC, SY+FS, SY+CC, FS+CC, and SY+FS+CC.
All conditions otherwise share the training setup.

We also evaluate a standing locomotion policy that tracks the same velocity commands.
This standing comparison condition is based on mjlab's \texttt{Mjlab-Velocity-Flat-Unitree-G1} and is trained without CC or SY; we remove torso linear velocity from its actor observation and use the same default arm posture as in the seated policies.

\begin{figure}[t]
    \centering
    \includegraphics[width=\linewidth]{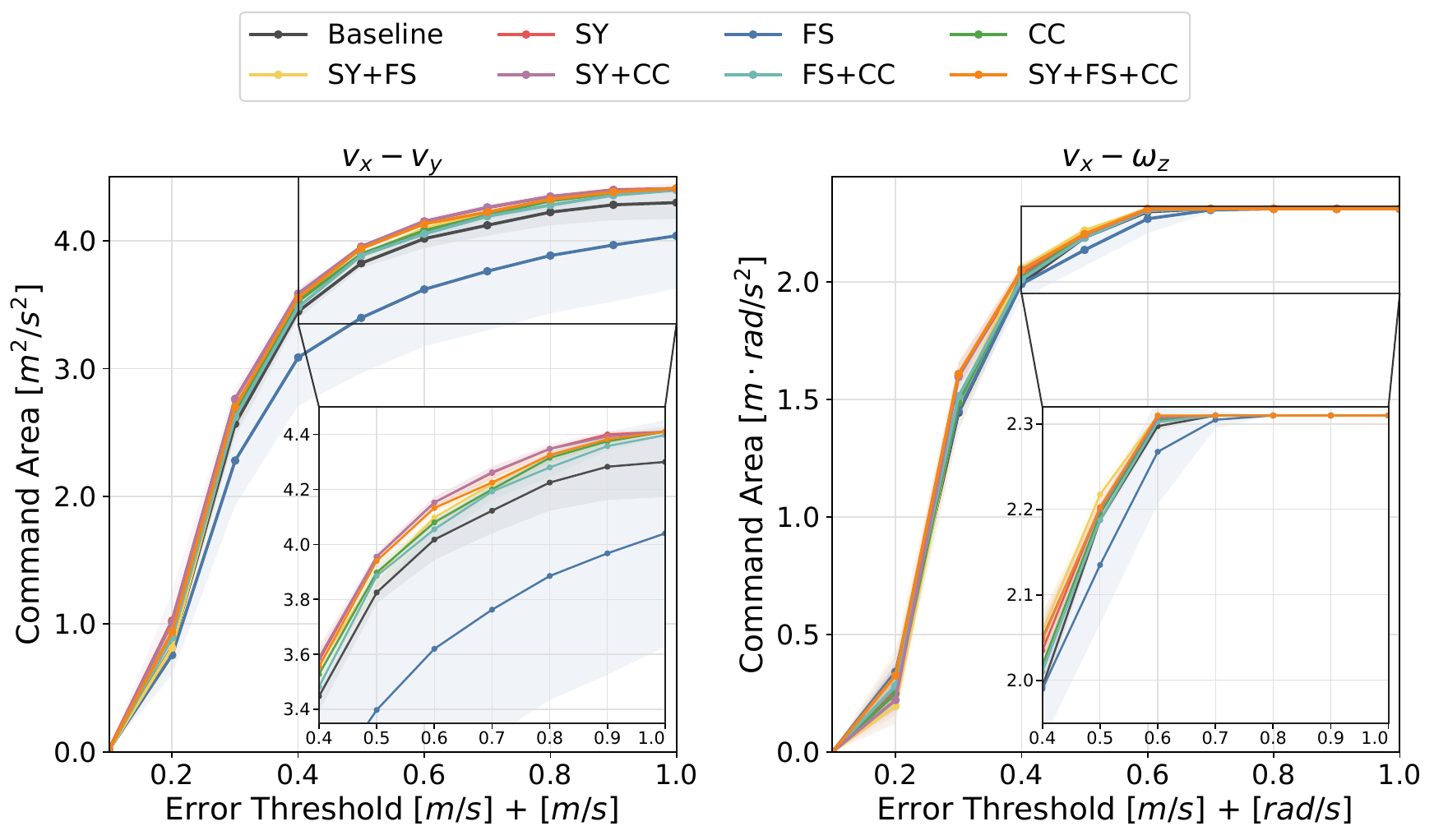}
    \caption{Command area as a function of the tracking-error threshold. Curves and shaded regions show the mean and standard deviation across four training seeds, respectively. }
    \label{fig:command_area}
\end{figure}

\section{Evaluation and Results}

\subsection{Overall Tracking and Training-Component Effects}
\label{chap:eval_random}

\begin{figure*}[!t]
    \centering
    \includegraphics[width=\linewidth]{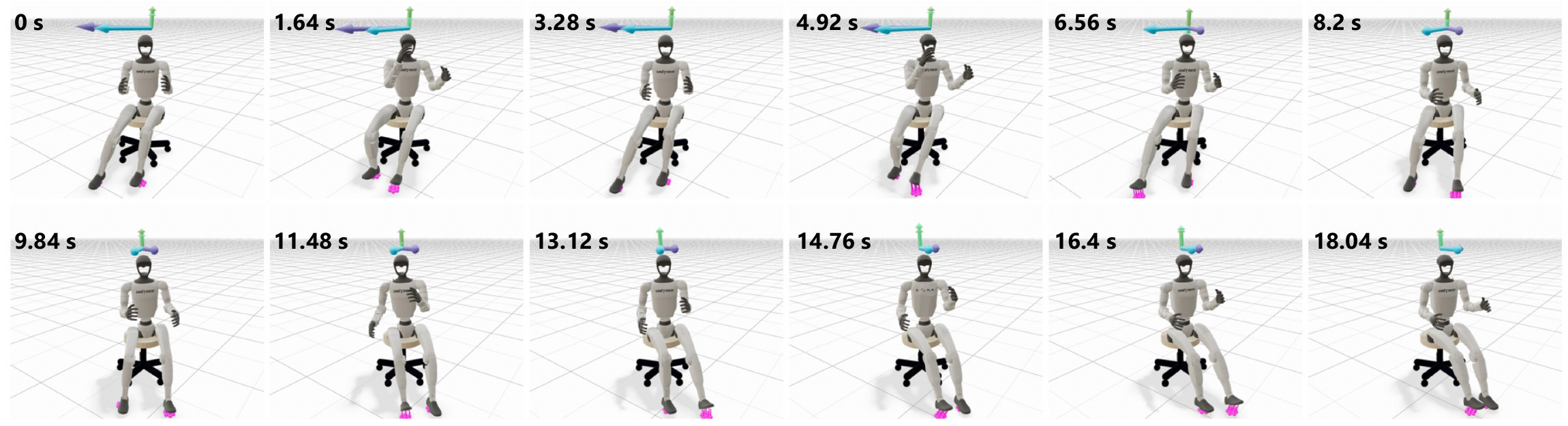}
    \caption{Example simulation rollout during a velocity-command transition from rightward to forward motion. Purple and cyan arrows indicate the commanded and measured planar velocities, respectively. The policy remained seated and adjusted its motion direction after the command switch.}
    \label{fig:sim_rollout}
\end{figure*}

\begin{table*}[!t]
    \centering
    \caption{Foot contact characteristics across movement directions.
        Contact fraction denotes left-only/right-only contact.
        Contact force denotes the resultant floor-contact force averaged during contact for the left/right foot.}
    \label{tab:direction_gait_results}
    \scriptsize
    \setlength{\tabcolsep}{1.8pt}
    \begin{subtable}[t]{0.485\textwidth}
        \centering
        \caption{Forward--backward}
        \begin{tabular}{@{}ll@{\hspace{6pt}}rr@{\hspace{6pt}}rr@{}}
            \toprule
                & & \multicolumn{2}{c}{Contact fraction [\%]}
                & \multicolumn{2}{c}{Contact force [N]} \\
            \cmidrule(lr){3-4}\cmidrule(l){5-6}
            Condition & Direction & L-only & R-only & L foot & R foot \\
            \midrule
            \multirow{2}{*}{CC}
                & Forward
                & $28.5\pm2.1$ & $23.5\pm2.3$
                & $156.8\pm4.4$ & $179.4\pm15.0$ \\
            & Backward
                & $40.1\pm4.0$ & $41.4\pm6.1$
                & $162.0\pm24.3$ & $153.9\pm27.1$ \\
            \midrule
            \multirow{2}{*}{SY+CC}
                & Forward
                & $25.1\pm1.5$ & $25.2\pm1.8$
                & $155.4\pm10.5$ & $156.1\pm11.0$ \\
            & Backward
                & $41.5\pm0.6$ & $41.5\pm0.8$
                & $167.1\pm6.6$ & $165.3\pm5.8$ \\
            \midrule
            \multirow{2}{*}{SY+FS+CC}
                & Forward
                & $27.4\pm0.7$ & $27.5\pm0.4$
                & $144.4\pm4.3$ & $145.6\pm4.9$ \\
            & Backward
                & $42.5\pm1.2$ & $41.8\pm1.2$
                & $168.4\pm4.8$ & $166.7\pm2.9$ \\
            \midrule
            \multirow{2}{*}{Standing}
                & Forward
                & $37.9\pm1.9$ & $36.1\pm2.3$
                & $274.5\pm1.8$ & $277.0\pm6.3$ \\
            & Backward
                & $41.5\pm0.9$ & $41.5\pm1.1$
                & $283.0\pm1.9$ & $281.0\pm2.5$ \\
            \bottomrule
        \end{tabular}
    \end{subtable}
    \hfill
    \begin{subtable}[t]{0.485\textwidth}
        \centering
        \caption{Leftward--rightward}
        \begin{tabular}{@{}ll@{\hspace{6pt}}rr@{\hspace{6pt}}rr@{}}
            \toprule
                & & \multicolumn{2}{c}{Contact fraction [\%]}
                & \multicolumn{2}{c}{Contact force [N]} \\
            \cmidrule(lr){3-4}\cmidrule(l){5-6}
            Condition & Direction & L-only & R-only & L foot & R foot \\
            \midrule
            \multirow{2}{*}{CC}
                & Leftward
                & $23.1\pm1.7$ & $57.3\pm0.7$
                & $86.1\pm3.6$ & $195.3\pm1.0$ \\
            & Rightward
                & $57.1\pm1.6$ & $20.8\pm1.5$
                & $202.6\pm5.1$ & $93.9\pm5.6$ \\
            \midrule
            \multirow{2}{*}{SY+CC}
                & Leftward
                & $19.2\pm1.2$ & $58.9\pm1.4$
                & $95.6\pm6.4$ & $193.6\pm4.3$ \\
            & Rightward
                & $58.7\pm1.2$ & $19.4\pm1.2$
                & $193.8\pm4.9$ & $94.6\pm6.4$ \\
            \midrule
            \multirow{2}{*}{SY+FS+CC}
                & Leftward
                & $21.4\pm1.6$ & $56.2\pm1.5$
                & $90.7\pm4.9$ & $202.9\pm4.7$ \\
            & Rightward
                & $56.3\pm1.6$ & $21.1\pm1.9$
                & $203.8\pm5.3$ & $91.8\pm5.3$ \\
            \midrule
            \multirow{2}{*}{Standing}
                & Leftward
                & $41.1\pm1.5$ & $45.1\pm2.0$
                & $299.1\pm6.3$ & $292.7\pm3.3$ \\
            & Rightward
                & $44.2\pm2.1$ & $43.0\pm2.5$
                & $296.3\pm3.9$ & $293.8\pm4.6$ \\
            \bottomrule
        \end{tabular}
    \end{subtable}
\end{table*}

Table~\ref{tab:random_command_results} reports the random-command evaluation results for the eight training conditions and Standing. Each policy was evaluated over 1,000 20-s rollouts, with the command resampled after 10~s and the training-time domain randomization, disturbances, and observation noise retained. The metrics included velocity-tracking RMSE, timeout success, seat-relative displacement, and torso tilt.

Across all eight training conditions, at least 99.45\% of the random-command rollouts were completed while maintaining small seat-relative displacement and torso tilt.
This high timeout success indicates maintenance of the seated state, but does not by itself demonstrate locomotion: a stationary policy can also avoid early termination.
Tracking RMSE and command area are therefore used to assess movement.
SY+CC achieved the lowest seed-averaged translational tracking errors and the highest timeout success rate, while SY+FS+CC gave similar tracking performance.
The best-performing training conditions were comparable to Standing, and both SY+CC and SY+FS+CC were numerically lower than Standing on all three tracking RMSEs. 
Because Standing was trained with a different environment and reward, these numbers contextualize the attainable tracking accuracy under the seated constraint rather than establishing superiority of either posture.

\begin{table*}[t]
    \centering
    \caption{Direction- and speed-dependent tracking, energy efficiency, and contact requirements.
            Contact fraction reports the mean single-foot/flight fraction, and contact force reports tangential/normal force on the floor.}
    \label{tab:speed_gait_results}
    \fontsize{6.5}{7.5}\selectfont
    \renewcommand{\arraystretch}{1.3}
    \setlength{\tabcolsep}{0.9pt}
    \begin{subtable}[t]{0.49\textwidth}
        \centering
        \caption{Forward--backward}
        \begin{tabular}{@{}lc|r@{\hspace{6pt}}r@{\hspace{6pt}}rr@{\hspace{4pt}}r@{}}
            \toprule
            Condition
                & \multicolumn{2}{c}{\makecell{Command\\{[m/s]}}}
                & \multicolumn{1}{c}{\makecell{Tracking RMSE\\{[m/s]}}}
                & \multicolumn{1}{c}{CoT}
                & \makecell{Contact fraction\\single [\%]\\flight [\%]}
                & \makecell{Contact force\\T [N]\\N [N]} \\
            \midrule
            \multirow[c]{4}{*}[-10pt]{SY+CC}
                & \multirow[c]{2}{*}{\rotatebox[origin=c]{90}{Forward}}
                & $+0.5$ & $0.039\pm0.008$ & $0.255\pm0.007$
                & \makecell[r]{$32.0\pm2.6$\\$36.1\pm5.3$}
                & \makecell[r]{$45.3\pm2.7$\\$80.8\pm3.0$} \\
            & & $+1.0$ & $0.308\pm0.004$ & $0.543\pm0.040$
                & \makecell[r]{$18.3\pm1.3$\\$63.3\pm2.6$}
                & \makecell[r]{$111.8\pm9.9$\\$188.0\pm16.1$} \\
            \addlinespace[2pt]
            & \multirow[c]{2}{*}{\rotatebox[origin=c]{90}{Backward}}
                & $-0.5$ & $0.033\pm0.015$ & $0.146\pm0.004$
                & \makecell[r]{$43.9\pm0.8$\\$0.0\pm0.0$}
                & \makecell[r]{$40.8\pm2.9$\\$109.1\pm6.7$} \\
            & & $-1.0$ & $0.132\pm0.020$ & $0.209\pm0.009$
                & \makecell[r]{$39.1\pm1.5$\\$21.6\pm3.1$}
                & \makecell[r]{$71.7\pm2.2$\\$201.1\pm6.6$} \\
            \midrule
            \multirow[c]{4}{*}[-10pt]{SY+FS+CC}
                & \multirow[c]{2}{*}{\rotatebox[origin=c]{90}{Forward}}
                & $+0.5$ & $0.042\pm0.008$ & $0.232\pm0.009$
                & \makecell[r]{$35.2\pm1.1$\\$29.6\pm2.2$}
                & \makecell[r]{$42.4\pm1.2$\\$78.1\pm3.7$} \\
            & & $+1.0$ & $0.321\pm0.009$ & $0.471\pm0.010$
                & \makecell[r]{$19.7\pm0.2$\\$60.6\pm0.4$}
                & \makecell[r]{$101.6\pm2.7$\\$173.3\pm5.2$} \\
            \addlinespace[2pt]
            & \multirow[c]{2}{*}{\rotatebox[origin=c]{90}{Backward}}
                & $-0.5$ & $0.037\pm0.022$ & $0.142\pm0.005$
                & \makecell[r]{$46.7\pm1.8$\\$0.1\pm0.1$}
                & \makecell[r]{$39.8\pm2.2$\\$109.9\pm5.8$} \\
            & & $-1.0$ & $0.092\pm0.030$ & $0.221\pm0.004$
                & \makecell[r]{$37.5\pm1.1$\\$25.0\pm2.3$}
                & \makecell[r]{$73.6\pm2.4$\\$203.2\pm7.9$} \\
            \bottomrule
        \end{tabular}
    \end{subtable}
    \hfill
    \begin{subtable}[t]{0.49\textwidth}
        \centering
        \caption{Leftward--rightward}
        \begin{tabular}{@{}lc|r@{\hspace{6pt}}r@{\hspace{6pt}}rr@{\hspace{4pt}}r@{}}
            \toprule
            Condition
                & \multicolumn{2}{c}{\makecell{Command\\{[m/s]}}}
                & \multicolumn{1}{c}{\makecell{Tracking RMSE\\{[m/s]}}}
                & \multicolumn{1}{c}{CoT}
                & \makecell{Contact fraction\\single [\%]\\flight [\%]}
                & \makecell{Contact force\\T [N]\\N [N]} \\
            \midrule
            \multirow[c]{4}{*}[-10pt]{SY+CC}
                & \multirow[c]{2}{*}{\rotatebox[origin=c]{90}{Leftward}}
                & $+0.5$ & $0.044\pm0.018$ & $0.231\pm0.004$
                & \makecell[r]{$40.0\pm2.7$\\$11.5\pm4.3$}
                & \makecell[r]{$48.4\pm1.3$\\$109.8\pm2.1$} \\
            & & $+1.0$ & $0.059\pm0.011$ & $0.273\pm0.006$
                & \makecell[r]{$38.0\pm1.2$\\$23.2\pm3.2$}
                & \makecell[r]{$72.7\pm4.4$\\$150.7\pm7.9$} \\
            \addlinespace[2pt]
            & \multirow[c]{2}{*}{\rotatebox[origin=c]{90}{Rightward}}
                & $-0.5$ & $0.042\pm0.017$ & $0.231\pm0.005$
                & \makecell[r]{$40.0\pm2.1$\\$11.3\pm3.7$}
                & \makecell[r]{$48.3\pm1.4$\\$109.5\pm2.5$} \\
            & & $-1.0$ & $0.059\pm0.013$ & $0.273\pm0.006$
                & \makecell[r]{$38.1\pm1.4$\\$23.1\pm3.6$}
                & \makecell[r]{$72.4\pm4.4$\\$150.3\pm8.3$} \\
            \midrule
            \multirow[c]{4}{*}[-10pt]{SY+FS+CC}
                & \multirow[c]{2}{*}{\rotatebox[origin=c]{90}{Leftward}}
                & $+0.5$ & $0.050\pm0.012$ & $0.227\pm0.008$
                & \makecell[r]{$40.0\pm1.6$\\$12.1\pm1.9$}
                & \makecell[r]{$48.9\pm2.0$\\$113.9\pm3.6$} \\
            & & $+1.0$ & $0.052\pm0.011$ & $0.264\pm0.008$
                & \makecell[r]{$37.6\pm1.8$\\$24.6\pm3.7$}
                & \makecell[r]{$72.3\pm2.9$\\$151.3\pm5.3$} \\
            \addlinespace[2pt]
            & \multirow[c]{2}{*}{\rotatebox[origin=c]{90}{Rightward}}
                & $-0.5$ & $0.051\pm0.012$ & $0.226\pm0.008$
                & \makecell[r]{$40.2\pm1.2$\\$12.0\pm1.5$}
                & \makecell[r]{$49.1\pm2.0$\\$114.4\pm3.6$} \\
            & & $-1.0$ & $0.052\pm0.013$ & $0.266\pm0.007$
                & \makecell[r]{$37.1\pm2.3$\\$25.7\pm4.6$}
                & \makecell[r]{$72.9\pm3.5$\\$152.6\pm5.9$} \\
            \bottomrule
        \end{tabular}
    \end{subtable}
\end{table*}

Figs.~\ref{fig:tracking_error_heatmap} and~\ref{fig:command_area} report the fixed-command grid evaluation, in which each command was held for 10~s and evaluated over 20 rollouts.
Following Margolis et al.~\cite{margolis2022}, the tracking error heatmap is computed as the sum of the axis-wise RMSEs in the \(v_x\)--\(v_y\) and \(v_x\)--\(\omega_z\) planes. For an error threshold \(\epsilon_0\), the command area is the total area of grid cells whose summed error is below \(\epsilon_0\). Fig.~\ref{fig:tracking_error_heatmap} shows tracking-error heatmaps for the four FS-containing training conditions, while Fig.~\ref{fig:command_area} shows command-area curves for all eight training conditions.

In the $v_x$--$v_y$ plane in Fig.~\ref{fig:command_area}, SY+CC and SY+FS+CC maintained large command areas across the error thresholds, whereas FS yielded a smaller command area and substantially larger variation across training seeds.
In the $v_x$--$\omega_z$ plane, the differences among conditions were smaller, and SY+FS attained the largest command area at low error thresholds.

The increased translational tracking errors under FS arose from extremely poor diagonal-forward tracking by some seeds, as shown in the left panel of Fig.~\ref{fig:tracking_error_heatmap}(a).
Inspection of the corresponding policy rollout showed that the robot had converged to a local optimum in which it remained completely stationary under diagonal-forward velocity commands.
In contrast, in SY+FS and FS+CC, even the worst-performing of the four seeds, shown in Fig.~\ref{fig:tracking_error_heatmap}(b) and (c), achieved lower tracking errors than the best FS seed shown in the right panel of Fig.~\ref{fig:tracking_error_heatmap}(a).
Across the four evaluated seeds, this stationary failure occurred only with FS alone and was avoided when FS was combined with either SY or CC.

Fig.~\ref{fig:sim_rollout} shows an example random-command rollout in which the policy remained seated and redirected the robot and chair after the command switched from rightward to forward motion.

\subsection{Direction- and Speed-Resolved Gait Characteristics}
Table~\ref{tab:direction_gait_results} reports direction-dependent foot-contact characteristics for CC, SY+CC, SY+FS+CC, and Standing. These conditions were selected to compare the effects of SY and FS under a common command curriculum, together with standing locomotion. Each policy was evaluated over 200 rollouts per command during forward, backward, leftward, and rightward motion at 0.5 and 1.0~m/s. The first 5~s served as warm-up, and the subsequent interval of up to 15~s was used for measurement. Domain randomization, velocity disturbances, and observation noise were disabled.

During longitudinal motion, CC produced left--right differences in contact fraction and force, whereas SY+CC and SY+FS+CC yielded nearly identical left and right values. 
Standing, trained without SY, also produced nearly balanced contacts.
In CC, the left--right differences and their across-seed variation were an order of magnitude larger than in SY+CC, yet tracking and seated stability were almost unchanged (Table~\ref{tab:random_command_results}).

During lateral motion, the leg opposite the movement direction had both the larger single-foot contact fraction and the larger contact force under all conditions, identifying it as the primary propulsive leg; the pattern reversed between leftward and rightward motion.

Table~\ref{tab:speed_gait_results} compares direction- and speed-dependent tracking, energy efficiency, contact states, and floor-contact forces for SY+CC and SY+FS+CC, the two best-performing training conditions in Sec.~\ref{chap:eval_random}. Energy efficiency was quantified using the cost of transport (CoT), defined as
\begin{equation}
    \mathrm{CoT}
    =
    \frac{W_{+}}{Mgd},
    \quad
    W_{+}
    =
    \sum_{t=1}^{K}
    \sum_m
    \max\left(\tau_{m,t}\dot{q}_{m,t},0\right)\Delta t ,
\end{equation}
where $\tau_{m,t}$ and $\dot{q}_{m,t}$ are the torque and velocity of joint $m$ at control step $t$, $\Delta t=0.02$~s is the control period, $K$ is the number of control steps in the measurement interval, $M$ is the robot mass, $g$ is gravitational acceleration, and $d$ is the horizontal torso-root path length over the measurement interval.

At 0.5~m/s, forward and lateral CoT values were relatively close.
At 1.0~m/s,  CoT followed backward $<$ lateral $\ll$ forward for both
SY+CC and SY+FS+CC, with forward CoT roughly twice the lateral value.
Fast forward motion also increased tracking RMSE, flight fraction, and contact forces.
SY+FS+CC reduced CoT relative to SY+CC for several fixed commands, particularly forward motion, whereas SY+CC generally gave lower tracking RMSE, indicating a trade-off between tracking and FS-induced motion regularization.

The rollouts showed that backward motion involved placing the foot flat in front of the body and then extending the knee while keeping the heel firmly planted, thereby propelling the robot and chair rearward. Lateral motion used a similar planted-leg extension pattern, with the propulsive leg located on the side opposite the direction of travel. Forward motion showed the opposite knee action: after making heel-first contact in front of the body, the robot flexed the knee while maintaining contact, moving itself and the chair toward the planted foot. Thus, backward and lateral motion were characterized by planted-leg extension, whereas forward motion involved heel-first contact followed by knee flexion. 

\begin{figure*}[t]
    \centering
    \begin{subfigure}{\textwidth}
        \centering
        \includegraphics[width=\linewidth]{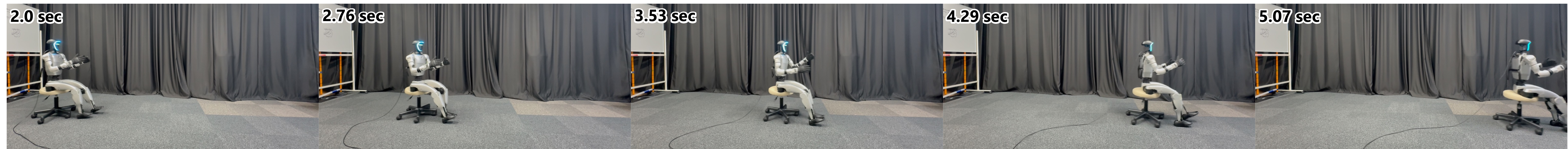}
        \caption{Forward}
        \label{fig:real_forward}
    \end{subfigure}
    \begin{subfigure}{\textwidth}
        \centering
        \includegraphics[width=\linewidth]{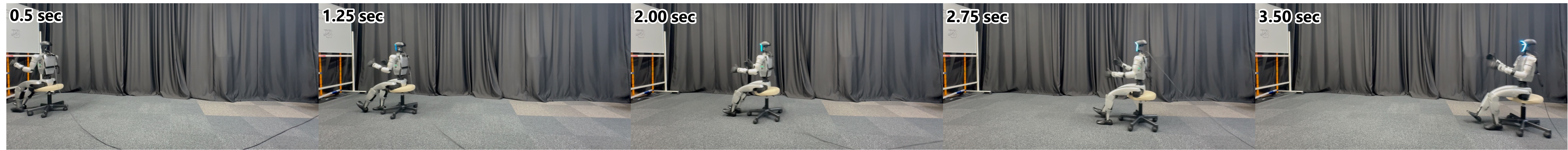}
        \caption{Backward}
        \label{fig:real_backward}
    \end{subfigure}
    \begin{subfigure}{\textwidth}
        \centering
        \includegraphics[width=\linewidth]{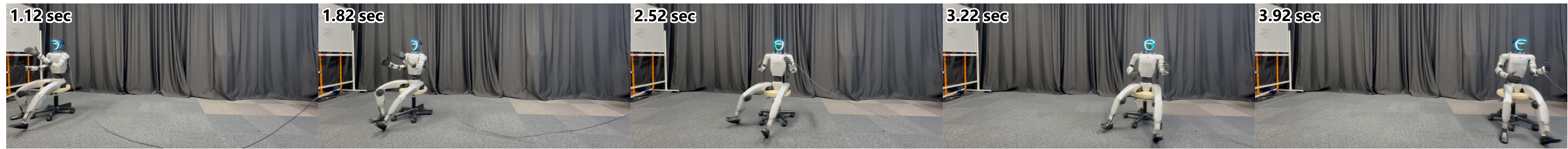}
        \caption{Leftward}
        \label{fig:real_leftward}
    \end{subfigure}
    \caption{Snapshots from sim-to-real deployment of the seated locomotion policy on a physical Unitree G1.}
    \label{fig:real_robot_directional_motion}
\end{figure*}

\subsection{Sim-to-Real Deployment}
To evaluate zero-shot sim-to-real transfer, we deployed the policy directly on a physical Unitree G1.
As shown in Figs.~\ref{fig:real_robot_teaser} and~\ref{fig:real_robot_directional_motion}, it qualitatively generated forward, backward, lateral, and turning motions while maintaining a seated posture, using only proprioception and velocity commands without contact sensing or chair states.
Tracking accuracy and disturbance robustness were not quantified on the physical robot.

\section{Discussion}

The results answer the question posed in Sec.~\ref{sec:intro}: a minimal extension of a standard standing velocity-tracking environment sufficed to learn seated locomotion, whose policies tracked omnidirectional velocity commands while completing nearly all 20-s random-command rollouts.
Because the actor, as in standing-locomotion policies, used only proprioception and velocity commands, the seated task added no sensing requirements for deployment, and the policy transferred zero-shot to a physical Unitree G1.

The training components introduced in this study---SY, FS, and CC---affected different aspects of the learned behavior.
FS, which regularized foot motion, reduced CoT while increasing tracking error, revealing a trade-off between energy efficiency and command-tracking performance.
This result indicates that excessively suppressing foot motion restricts not only unnecessary motion but also the propulsive motion required for velocity tracking.
Tuning the strength of this regularization may be more difficult for seated locomotion than for standing locomotion.
In the FS-only condition, excessive regularization caused some seeds to converge to a stationary local optimum that ignored diagonal-forward velocity commands.
When either SY or CC was added to FS, the evaluated seeds did not converge to the same local optimum, even without retuning the FS weight.
With SY, symmetry augmentation increased the number of samples used for training, potentially reducing the relative influence of samples that induced the local optimum.
With CC, the required foot-trajectory patterns were first learned within a low-speed command range, where smaller foot forces suffice to propel the robot and chair without slipping. Expanding the command range only afterward may have helped avoid the stationary solution.
Verifying this explanation would require testing a range of curriculum schedules, such as different initial ranges and expansion timings, which we leave for future work.

SY also reduced the left--right bias in leg use during longitudinal motion. In standing locomotion, the legs support the body, so biased leg use compromises posture and is corrected during training, as observed in the Standing condition. In seated locomotion, the chair supports the body, and the biased CC policies maintained the seated state and tracked commands almost as well as SY+CC, so nothing in the task corrects the bias, and its direction varied across seeds. Symmetric leg use remains desirable for avoiding persistent unilateral loading and for predictable responses to mirrored commands, and SY provided it. A possible source of the initial bias is the unfixed pelvis--seat configuration, which may favor one leg for propulsion depending on how the robot settles on the seat; we did not measure this and leave it for future work.

Regarding the direction dependence of CoT, Table~\ref{tab:speed_gait_results}(a) shows that forward and backward tangential forces were comparable at low speed, but at high speed, forward motion exhibited greater tangential force despite lower normal force. The nearly speed-invariant \(T/N\) within each direction suggests that the direction-specific contact-loading pattern was retained as speed increased. Fast forward locomotion may therefore have operated closer to the friction limit \(T\leq\mu N\), where $\mu$ is the friction coefficient, increasing susceptibility to foot slip and reducing the efficiency with which joint work was converted into propulsion. This may partly explain its higher CoT and tracking error.

The Appendix reports a preliminary stationary power comparison: mean battery-side power was 111.06~W while standing and 101.25~W while seated, equivalent to approximately 22~min longer operation under a nominal full-capacity extrapolation.
Because only one trial per posture was measured, and actuator-level power and heat were not isolated, this result does not establish a general energy-saving benefit.

The study is limited to one chair, floor condition, and command range.
In simulation, the chair was modeled with spherical casters and a rigid seat; the effects of caster steering dynamics, seat compliance, and seat height on the learned contact patterns and energy efficiency remain to be examined.

Future work will integrate reaching and pick-and-place for desk-to-desk seated loco-manipulation, investigate energy-aware direction selection, and develop an impulse-and-coast strategy in which the chair coasts after a propulsive impulse.
The flight fraction exceeding $60\%$ under the 1.0-m/s forward command (Table~\ref{tab:speed_gait_results}(a)) suggests that the current behavior could be developed toward such a strategy.

\section{Conclusion}
In this study, we formulated seated locomotion on a passive mobile chair with casters as omnidirectional velocity tracking with unfixed pelvis--seat contact and intermittent foot--floor propulsion. We showed that a minimal extension of a standard standing velocity-tracking environment, consisting of a passive-chair model, seated-state rewards, critic-only chair observations, and pelvis--seat contact settings, suffices to learn this task without motion references. The learned policies tracked omnidirectional commands through nearly all 20-s rollouts with accuracy comparable to standing locomotion, and the actor, using only proprioception and velocity commands, transferred zero-shot to a physical Unitree G1. Two training components proved important, likely because the chair supports the body. A command curriculum (CC) that learns low-speed propulsion before expanding the command range avoided stationary local optima under diagonal-forward commands, and symmetry regularization (SY) reduced the left--right bias in leg use that otherwise persisted. Direction-resolved analysis further revealed that propulsion patterns and cost of transport differ markedly across movement directions, with forward motion being the most costly. These findings provide a basis for future work toward seated loco-manipulation.

\appendix[Stationary Power Comparison]
\label{app:stationary_power}

\begin{figure}[!ht]
    \centering
    \includegraphics[width=0.85\columnwidth]{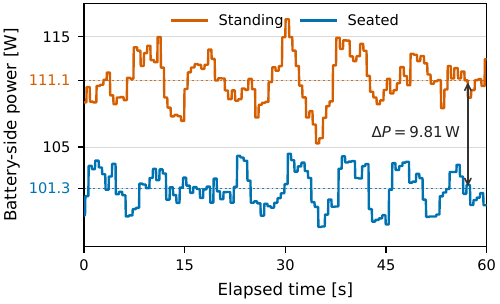}
    \caption{Battery-side power; dotted lines show means.}
    \label{fig:bms_power}
\end{figure}

From one 60~s trial per posture, mean battery-side power was 111.06~W while standing and 101.25~W while seated (Fig.~\ref{fig:bms_power}), a 9.81~W difference. For the 421.2~Wh battery, constant-power extrapolation gives 3.79 and 4.16 h, respectively (+22 min).

\bibliographystyle{IEEEtran}
\bibliography{references}

\end{document}